# Understanding Driver Cognition and Decision-Making Behaviors in High-Risk Scenarios: A Drift Diffusion Perspective

Heye Huang[a], Zheng Li[a *], Hao Cheng[b], Haoran Wang[c], Junkai Jiang[b], Xiaopeng Li[a *], Arkady Zgonnikov[d]

[a] Department of Civil and Environmental Engineering, University of Wisconsin-Madison, WI 53706, USA
[b] School of Vehicle and Mobility, Tsinghua University, Beijing 100084, China
[c] Key Laboratory of Road and Traffic Engineering of the Ministry of Education, Tongji University, Shanghai 200092, China
[d] Department of Cognitive Robotics, Delft University of Technology, Delft 2628, The Netherlands

[*] Correspondence
E-mail address: hhuang468@wisc.edu (H. Huang), zli2674@wisc.edu (Z. Li), chenh22@mails.tsinghua.edu.cn (H. Cheng), wang_haoran@tongji.edu.cn (H. Wang), jiangjk21@mails.tsinghua.edu.cn (J. Jiang), xli2485@wisc.edu (X. Li), A.Zgonnikov@tudelft.nl (A. Zgonnikov).

**Abstract**

Ensuring safe interactions between autonomous vehicles (AVs) and human drivers in mixed traffic systems remains a major challenge, particularly in complex, high-risk scenarios. This paper presents a cognition-decision framework that integrates individual variability and commonalities in driver behavior to quantify risk cognition and model dynamic decision-making. First, a risk sensitivity model based on a multivariate Gaussian distribution is developed to characterize individual differences in risk cognition. Then, a cognitive decision-making model based on the drift diffusion model (DDM) is introduced to capture common decision-making mechanisms in high-risk environments. The DDM dynamically adjusts decision thresholds by integrating initial bias, drift rate, and boundary parameters, adapting to variations in speed, relative distance, and risk sensitivity to reflect diverse driving styles and risk preferences. By simulating high-risk scenarios with lateral, longitudinal, and multidimensional risk sources in a driving simulator, the proposed model accurately predicts cognitive responses and decision behaviors during emergency maneuvers. Specifically, by incorporating driver-specific risk sensitivity, the model enables dynamic adjustments of key DDM parameters, allowing for personalized decision-making representations in diverse scenarios. Comparative analysis with IDM, Gipps, and MOBIL demonstrates that DDM more precisely captures human cognitive processes and adaptive decision-making in high-risk scenarios. These findings provide a theoretical basis for modeling human driving behavior and offer critical insights for enhancing AV-human interaction in real-world traffic environments.

**Keywords**：Driver risk cognition, Driving simulator study, Drift diffusion model, Risk sensitivity

# 1. Introduction

Driving safety is directly influenced by drivers' risk cognition and collision avoidance decision-making abilities in high-risk scenarios. In real-world driving, risk cognition generally involves complex interactions among multiple co-existing risk factors rather than being limited to a single risk source (Crosato et al., 2024; Huang et al., 2022). In such complex scenarios, drivers are required to rapidly identify risks and take appropriate actions. However, their cognitive processes are intricate, involving multiple stages of cognition, decision-making, and control, making it difficult to quantify their risk responses and proactive behaviors (Aven, 2011; Wang et al., 2020). Therefore, it is essential to investigate how drivers dynamically utilize their risk cognition to make socially acceptable decisions in high-risk interactive environments (Zgonnikov et al., 2022).

Considerable attention has been directed toward understanding drivers' risk cognition and decision-making capabilities, leading to the development of numerous models such as car-following and lane-changing models (Amditis et al., 2010; Orfanou et al., 2022). These models are generally classified into five categories (Wang et al., 2022): behavioral simulation models, game-theoretic reasoning models, social force-driven two-dimensional models, learning-driven models, and information cognition models.

**Behavior simulation models** simulate drivers' actions under specific conditions. Advanced examples include micro-binary models, cellular automata models, and driver preview-follow models, implemented in simulators such as Simulation of Urban Mobility (SUMO) and public transport simulators like VISSIM, alongside lane-changing trajectory models derived from traffic flow theory (Fernandes and Nunes, 2010; Xu et al., 2012). Simplified dynamic behavior models, such as the intelligent driver model (IDM) (Derbel et al., 2013; Treiber et al., 2006) and the minimize overall braking induced by lane change model (MOBIL) (Kesting et al., 2007), account for fundamental vehicle dynamics and constitute single stimulus-response representations. These behavioral simulation approaches are accessible and widely utilized to reproduce diverse driving scenarios, effectively capturing both micro-scale individual behaviors and macro-scale group dynamics (Treiber and Kesting, 2017). However, such models exhibit notable limitations, primarily responding to unidimensional risk stimuli. A single mathematical framework often fails to comprehensively reflect or address the dynamic requirements and expectations of drivers. Moreover, behavioral simulation models do not adequately respond in real-time to evolving driver states nor elucidate underlying mechanisms driving behaviors in complex scenarios. For instance, scenarios involving interactions among multiple road users necessitate consideration of multidimensional risks, including lateral factors, road geometry, and infrastructure configurations.

**Game game-theoretic reasoning models** provide a rigorous approach to examining strategic interactions among rational agents, where the actions of each participant influence others' outcomes (Huo et al., 2023). Within game theory-based driver cognition modeling, dynamically integrating vehicle safety and comfort considerations with predictions of surrounding vehicles' intentions and behaviors is essential. Consequently, driver decisions emerge from multi-agent strategic interactions, resulting in optimal decision-making strategies. Bayesian dynamic models have similarly been employed for behavior inference (Schulz, 2021). For example, Darius et al. (Schulz et al., 2019) developed a probabilistic prediction framework utilizing dynamic Bayesian networks (DBN), explicitly modeling multi-vehicle interactions and employing context-aware motion representations to characterize driver behavior at intersections. By integrating prior knowledge and observational data, these game-theoretic approaches enable probabilistic prediction of driver actions (Huang et al., 2024). However, these models' reliance on assumptions of rational behavior limits their applicability to real-world scenarios, and they often inadequately capture the complexities of drivers' risk cognition. Moreover, decision-making complexity escalates significantly with the increase in involved agents and potential strategies.

**Social force-driven two-dimensional driver models** characterize driver responses to risk stimuli by incorporating social and virtual forces, enabling simultaneous quantification of longitudinal and lateral risks across various driving scenarios (Rosenberg, 1990). These models simulate human interactions under complex motion stimuli, grounded in the concept of driver risk cognition as influenced by virtual forces (Bieleke et al., 2020). Herbing et al. proposed a social force-evolutionary framework, employing virtual forces to elucidate social interactions, thereby enhancing human-like behaviors in multi-agent vehicle environments through reward-based design mechanisms (Helbing and Molnár, 1995). Similarly, David et al. introduced a two-dimensional driver risk field model capturing driver perceptions of event probabilities and providing quantitative assessments of perceived risks (Kolekar et al., 2020). In contrast to conventional dynamic behavior simulation methods, social force-based approaches effectively represent the multidimensionality of driver risks and explicitly address social interaction influences. Nevertheless, such force-based representations might oversimplify the intricate cognitive processes underlying individual driver decision-making.

**Learning-driven models** employ advanced neural network architectures, such as deep neural networks (DNN) and convolutional neural networks (CNN), to mine extensive driving datasets and extract intricate behavioral patterns (Li et al., 2022). For example, Sharifzadeh et al. (Sharifzadeh et al., 2016) employed deep Q-networks for deep reinforcement learning to investigate lane-changing and overtaking behaviors on highways. However, their study did not address vehicle safety and relied on simplified simulation scenarios. Such models excel at processing complex, high-dimensional data, enabling the identification of nuanced patterns within large-scale datasets (Kuutti et al., 2021; Schulte et al., 2022). Despite their effectiveness, learning-driven models often act as "black boxes," limiting interpretability and impeding the exploration of underlying cognitive and decision-making processes. This opacity poses significant challenges for practical application, particularly in safety-critical contexts where transparent and understandable decision-making mechanisms are essential. Consequently, enhancing interpretability in learning-driven approaches remains a crucial direction for future research aimed at improving our understanding of driver risk cognition and behavior.

**Information cognition models** leverage principles from cognitive psychology and neuroscience to simulate human perceptual processing, cognition, and decision-making mechanisms (Markkula et al., 2023). By modeling cognitive functions such as attention, memory, and learning, these models help understand and predict human behavior (Ratcliff et al., 2016). Notably, drift diffusion models (DDM) have significantly contributed to the understanding of psychological and neural underpinnings of driver decision-making. These models quantitatively represent decision processes as the gradual accumulation of information until a decision threshold is reached (Mohammad et al., 2024). Arkady et al. (Mohammad et al., 2023) integrated DDM-based decision thresholds to model driver judgments during unprotected left-turn maneuvers, effectively bridging cognitive theory and empirical driving research. Their approach provided real-time predictions regarding drivers' gap acceptance, closely reflecting natural cognitive processes. By explicitly simulating human cognitive mechanisms, information cognition models not only clarify the intricacies of driver risk cognition and decision-making but also closely align with observed behaviors (Siebinga et al., 2024; Zgonnikov et al., 2024). Consequently, these models are invaluable for the development of automated systems such as AVs, where a comprehensive understanding of human decision-making can significantly enhance both safety and operational performance.

Therefore, this paper aims to delve into drivers' risk cognition and decision-making processes in high-risk scenarios. From the nature of human behavior, we characterize and predict their decision-making behaviors during emergency evasive maneuvers. This understanding will provide insights for intelligent vehicles to tailor their interactive behaviors to adapt to complex and dynamic traffic environments. Our contributions are as follows:

- We introduce a cognitive model that employs the DDM. This model incorporates parameters such as initial decision bias, drift rate, and boundary separation based on the driver's speed and proximity to other road users, offering a quantitative framework to simulate human decision-making in high-risk scenarios.

- We develop a driver risk-sensitivity model based on a multivariate Gaussian distribution, which quantifies individual differences in drivers' risk cognition and is utilized to tailor personalized decision-making processes.

- We validate the models by simulating high-risk scenarios on a driving simulator platform. Extensive driving simulator experiment data demonstrates the model's effectiveness in predicting human drivers' cognitive processes and decisions under critical conditions. This empirical validation confirms the model's applicability to real-world driving environments.

The remainder of this paper is organized as follows: Section 2 introduces the experimental methodology employed for collecting driver behavior data via a driving simulator. Section 3 describes the formulation of the risk sensitivity model, capturing individual differences in driver risk perception. Section 4 elaborates on the modeling approach for human decision-making behavior. Sections 5 and 6 present the evaluation results and conclusions of the study, respectively.

## 2. High-risk scenario driver cognition dataset

In this section, a series of driving simulator experiments were conducted to examine human cognition and decision-making in high-risk scenarios. Participants engaged in simulated environments featuring diverse risk factors, such as sudden obstacles, unpredictable traffic flows, and high-density conditions.

*2.1 Participants*

A total of 58 licensed drivers with normal or corrected-to-normal vision participated in the experiment. Before testing, they completed a demographic and subjective questionnaire (Table 1) covering age, driving experience, annual mileage, accident history, and self-reported driving style (cautious, normal, or aggressive). Data from 58 participants (mean age = 36.5 years; standard deviation [SD] =8.37; range=22-55; 8 females and 50 males) were analyzed.

**Table 1.** Demographic variables for collected drivers.

|      | Age /*Year* | Driving years/*Year* | Average driving time/*Hour* | Mileage/ *Kilometer* |
|------|-------------|----------------------|------------------------------|----------------------|
| Mean | 36.50       | 12.10                | 39.29                        | 23731.43             |
| SD   | 8.37        | 7.10                 | 38.24                        | 19213.04             |

*2.2 Experimental platform*

The driving simulation platform is extensively used to evaluate the impact of driving performance on conflict risk. As illustrated in Fig. 1, the simulator hardware consists of a Logitech G29 steering wheel, accelerator and brake pedals, and display systems for the driving environment. It features a linear motion base with one degree of freedom (pitch) and a full-scale cabin equipped with a realistic operation interface, ambient noise simulation, and motion feedback, along with digital video playback and vehicle dynamics modeling. The simulation environment provides a 300-degree field of view at a resolution of 1400 × 1050 pixels, including left, center, and right rearview mirrors. The supporting software enables customized scenario design, virtual traffic environment simulation, and road modeling, facilitating road construction, traffic flow generation, and traffic control.

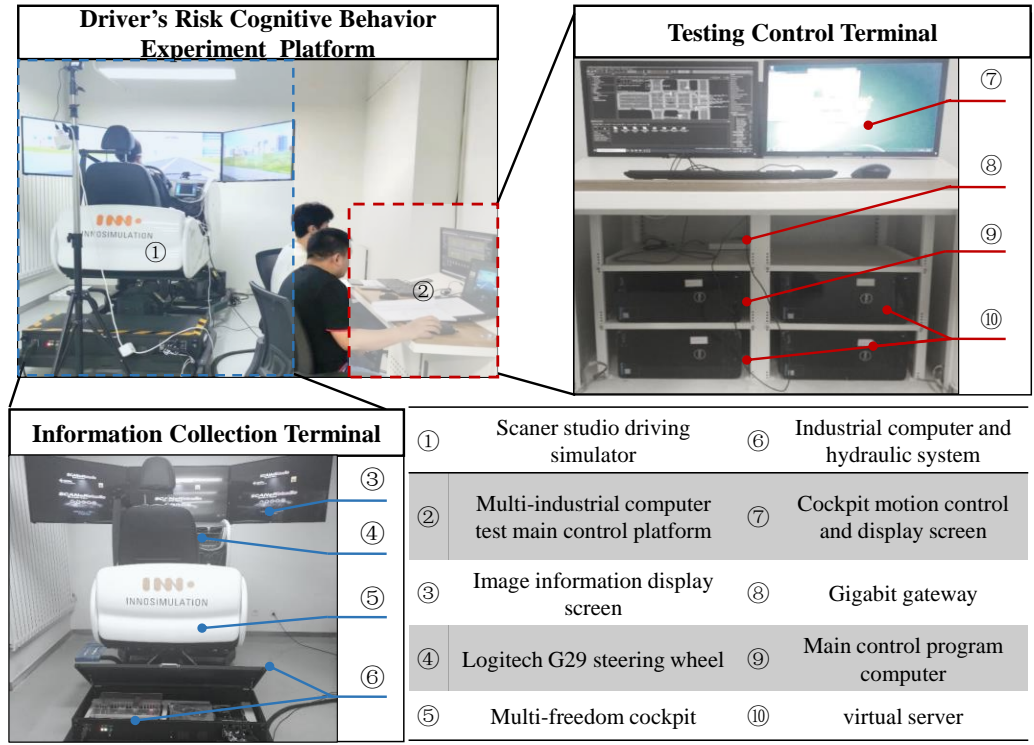

**Figure 1. Driving simulator platform.**

*2.3 Experimental design and analysis*

To address conflicts among vehicles, environmental factors, and road users, our design is informed by accident statistics from NHTSA and GES, emphasizing rear-end collisions (29%), intersection crossings (24%), road departures (19%), and lane changes (12%) (Wang et al., 2020). The experimental scheme incorporates diverse risk sources, varying stimulus timing conditions (4 seconds vs. 8 seconds), and three scenario types: cut-in (lateral risk), rear-end collision (longitudinal risk), and lane-changing (multi-dimensional risk), as illustrated in Fig. 2.

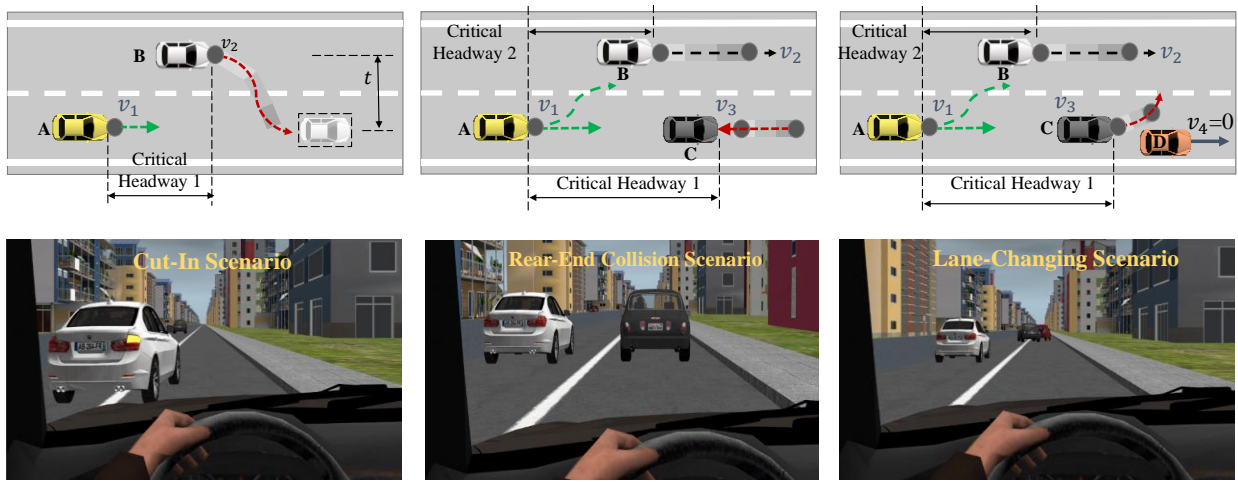

**Figure 2. The experimental scenarios.**

The experimental scenario setting details are as follows. Specifically, in the cut-in scenario (lateral risk source), we consider two vehicles: the ego vehicle A and the surrounding vehicle B. A accelerates to 80-120 km/h in the right lane, while B, 20 meters away, suddenly cuts in at 120 km/h, forcing A to perform emergency evasive actions. This scenario concludes within 5-6 seconds after the maneuver. In the rear-end collision scenario (longitudinal risk source), we involve three vehicles: the ego vehicle A, surrounding vehicle B, and lead vehicle C. Specifically, vehicle A, driven by a human, and vehicles B and C, set to a constant speed of 80 km/h by the simulator, are tested. When C abruptly brakes at -8 m/s², A must decide quickly whether to change lanes or stop abruptly, as outlined in Table 2. In the lane-changing scenario (multiple risk sources), the speed details remain consistent. We consider four vehicles: A, B, C and a stationary vehicle D. With a static obstacle D ahead, C suddenly changes lanes, requiring A's driver to also engage in interactive driving. Notably, throughout the experiment, the test driver of A was unaware of the behavior settings of surrounding vehicles, allowing for a more authentic capture of reaction time, deceleration, and other behavior characteristics under unexpected conditions.

As depicted in Fig. 3, the multi-stage driver behavior experiment, based on the risk cognition experimental paradigm, consists of three key phases: (1) Pre-experiment preparation: Includes driver information collection, eye tracker calibration, and scenario familiarization. (2) In-lab testing: A structured driving process where drivers transition from normal driving to stimulus onset, followed by risk response and scenario conclusion. Fig. 3 illustrates these steps in a stimulus-response time window, detailing the transition from normal driving (4s) to risk stimulus (4s), followed by the driver's response (2s). (3) Post-experiment processing: Extracts reaction time, steering/braking behavior, and acceleration/deceleration responses via video playback and feature quantification. This structured design ensures naturalistic driver behavior data collection, enabling a quantitative evaluation of decision-making and evasive maneuvers across different risk scenarios.

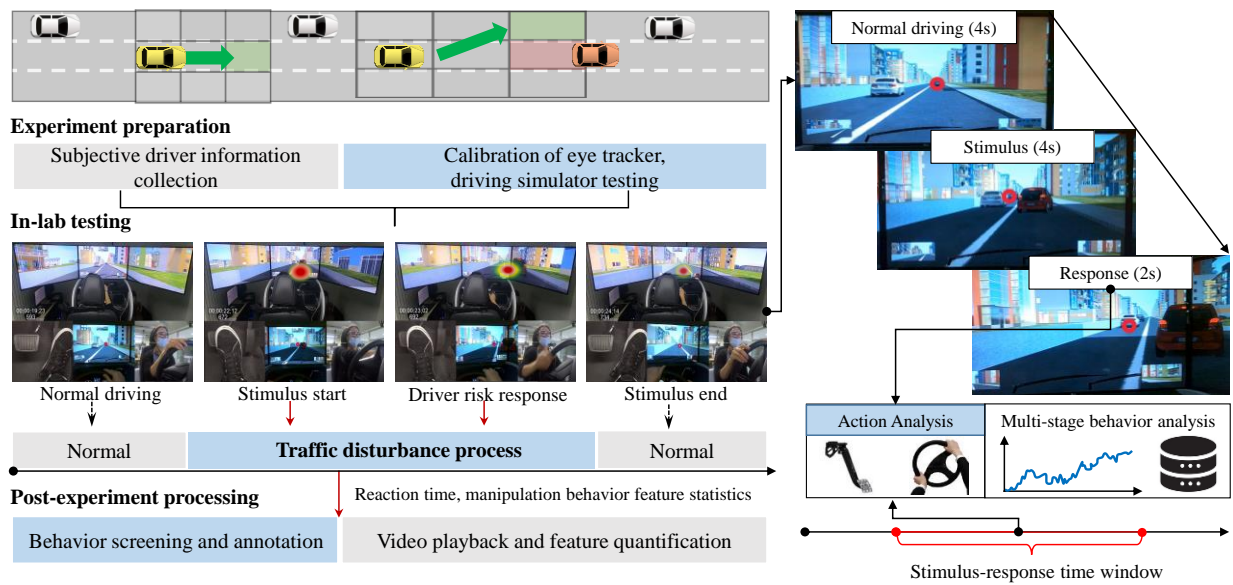

**Figure 3. The multi-stage driver behavior experiment paradigm.**

*2.4 Data processing and key variable extraction*

In the cognition process, key variables were extracted to assess driver reactions in high-risk scenarios. Cognitive Reaction Time (CRT) measures the time from hazard appearance to driver recognition, determined via eye-tracking fixation or initial control input (braking/steering). Brake Reaction Time (BRT) quantifies the delay from a traffic disturbance to brake activation, reflecting emergency braking responsiveness. Speed Adjustment Time (SAT) captures the interval from brake initiation to maximum deceleration, indicating driver deceleration strategies. For the decision process, we extracted variables related to braking and steering strategies. Maximum deceleration evaluates braking intensity, while Minimum Time to Collision (TTC) indicates collision risk, with lower values signifying higher danger levels. Maximum steering angle assesses evasive maneuvering, where larger angles indicate more aggressive avoidance strategies.

**Table 2.** Parameters of decision-making and control behavior in high-risk scenarios.

| Parameters | High-risk scenarios | | |
| --- | --- | --- | --- |
| | Cut-in | Rear-end collision | Lane-changing |
| Frequency | 58 | 58 | 58 |
| Disturbance occurrence (s) | 22.54±0.72 | 21.20±1.93 | 5.47±0.68 |
| Brake reaction time (s) | 1.62±0.33 | 1.42±0.35 | 0.73±0.48 |
| Speed adjustment time (s) | 0.73±0.26 | 0.64±0.35 | 1.58±0.74 |
| Maximum deceleration (m/s$^2$) | -8.39±1.52 | -7.92±2.33 | -9.24±2.01 |
| Minimum time to collision (s) | 0.92±0.12 | 0.52±0.14 | 0.74±0.47 |
| Braking spatial distance (m) | 8.35±1.21 | 42.15±7.62 | 31.77±7.46 |
| Maximum steering angle (°) | 32.14±9.40 | 36.74±9.40 | 39.96±12.47 |
| Collision avoidance measures | Brake/Steer (54/4) | Brake/Steer (9/49) | Brake/Steer (12/46) |
| Number of accidents | 2 | 18 | 12 |

Table 2 summarizes decision-making and control behavior parameters across different high-risk scenarios. Scenarios like cut-ins and left turns show larger minimum TTCs (0.92±0.12s, 1.04±0.26s), indicating lower risks, and lower maximum decelerations in left-turn scenarios (-7.74±1.27s) suggest gentler braking upon disruption detection. Fig. 4 illustrates notable differences in driver decisions under identical conditions, influenced by cognitive response times, which impact collision or safety outcomes. These visual analyses enhance the understanding of individual risk cognition and response strategies.

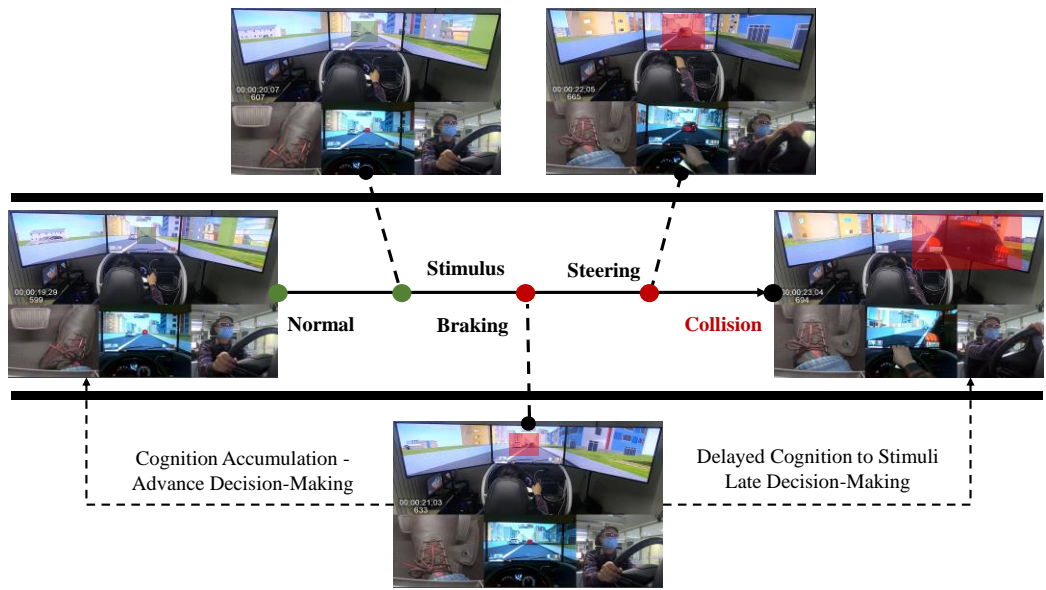

**Figure 4. Spatiotemporal distribution of decision-making behaviors in critical scenarios.**

## 3. Driver risk sensitivity modeling

Driver risk sensitivity characterizes an individual's response to uncertain environments and potential risks, influencing their decision-making and collision avoidance strategies. In this study, risk sensitivity is quantified using vehicle kinematics, focusing on both longitudinal and lateral dynamics. To differentiate risk sensitivity levels among drivers and across scenarios, we define three sub-models representing high, medium, and low sensitivity levels: $R_s = \{R_{s,h}, R_{s,m}, R_{s,l}\}$. These sub-models are derived from braking initiation velocity $v_b$, longitudinal acceleration $a_x$, and lateral acceleration $a_y$, which reflect a driver's evasive response intensity—higher acceleration magnitudes indicate greater sensitivity and more aggressive actions.

*3.1 Driver risk sensitivity with multivariate Gaussian distribution*

To model driver risk sensitivity, we employ a Multivariate Gaussian Distribution (MGD), which effectively captures variations in risk cognition and decision-making tendencies. This probabilistic framework characterizes driver behavior by representing the distribution of collision avoidance strategies under identical high-risk scenarios.

In driver risk sensitivity modeling, selecting an appropriate probability density function (PDF) is essential for accurately capturing decision-making patterns. Skewness and kurtosis, as key statistical measures, allow for a quantitative evaluation of decision tendencies. Due to its robustness in modeling behavioral variability, MGD serves as a reliable framework for describing risk sensitivity, enabling the differentiation between common decision patterns across drivers and individualized risk-sensitive behaviors. For a bivariate random variable (e.g., longitudinal acceleration $a_y$ and lateral acceleration $a_x$ ), the probability density function is given by:

$$f(x) = \frac{1}{\sqrt{(2\pi)^m |\Sigma|}} e^{-\frac{1}{2}(x-\mu)^T \Sigma^{-1}(x-\mu)} \tag{1}$$

where, $|\Sigma|$ represents the determinant of the covariance matrix, and $\mu = E[x]$ denotes the mean of the random variable $x$.

Next, we use maximum likelihood estimation to fit the multivariate Gaussian model to driving behavior data, estimating parameters for precise distribution fitting. The likelihood function $L(\mu, \Sigma)$ for $n$ samples is defined in Eq. (2). By deriving and setting the partial derivatives of $\mu$ and $\Sigma$ to zero, we obtain the estimates $\hat{\mu}$ and $\hat{\Sigma}$. These parameter estimates are detailed in Eqs. (3) and (4).

$$\begin{aligned} L(\mu, \Sigma) &= \prod_{i=1}^{n} f(x_i; \mu, \Sigma) \\ &= \prod_{i=1}^{n} (2\pi)^{-\frac{m}{2}} |\Sigma|^{-\frac{1}{2}} \exp\left(-\frac{1}{2}(x_i - \mu)^T \Sigma^{-1}(x_i - \mu)\right) \\ &= (2\pi)^{-\frac{nm}{2}} |\Sigma|^{-\frac{n}{2}} \exp\left(-\frac{1}{2}\sum_{i=1}^{n}(x_i - \mu)^T \Sigma^{-1}(x_i - \mu)\right) \end{aligned} \tag{2}$$

$$\hat{\mu} = \bar{x} \tag{3}$$

$$\hat{\Sigma} = \frac{1}{N} \sum_{i=1}^{n} (x_i - \bar{x})(x_i - \bar{x})^T \tag{4}$$

Finally, the distribution characteristics of control behavior parameters for different drivers in the same scenario can be accurately represented using the multivariate Gaussian model.

*3.2 Model performance analysis*

Fig. 5 displays the driver risk sensitivity model results across various urgent scenarios, including cut-in, rear-end collision, and lane-changing high-risk scenarios, capturing variations in driving behavior. (1) Cut-in scenario ($R_s$ impact on longitudinal deceleration): A narrow, high-peaked distribution indicates consistent driver responses, primarily relying on longitudinal deceleration with minimal lateral movement. (2) Rear-end collision scenario ($R_s$ impact on braking intensity): A broader distribution suggests greater variability, with some drivers applying strong braking while others adopt gradual deceleration. (3) Lane-changing scenario ($R_s$ impact on maneuver selection): A flatter, more dispersed distribution reflects higher variation in longitudinal and lateral accelerations, indicating a mix of steering and braking maneuvers.

These results align with the expected behavioral patterns: (1) Higher acceleration in high-risk scenarios: As risk levels increase, drivers exhibit stronger braking and steering responses, resulting in higher longitudinal ($a_x$) and lateral ($a_y$) accelerations. This is particularly evident in rear-end collision and lane-changing scenarios, where probability distributions are broader, indicating more intense reactions. (2) Significant individual differences in risk sensitivity: Some drivers consistently demonstrate higher acceleration values, reflecting a more aggressive evasive strategy, while others show lower response magnitudes, favoring a more conservative avoidance approach. The results confirm that MGD-based risk sensitivity modeling effectively quantifies variations in driver decision-making.

These findings are consistent with existing literature and Gaussian-based driver behavior models (Zhou and Zhong, 2020), while our risk sensitivity model provides a more refined characterization, accurately distinguishing behavioral differences across scenarios. Additionally, it predicts probability distributions under the same urgency level, segmenting sub-models for different driver responses in high-risk conditions. This enhances the quantification of decision-making tendencies, supporting driving behavior prediction, risk assessment, and adaptive autonomous control.

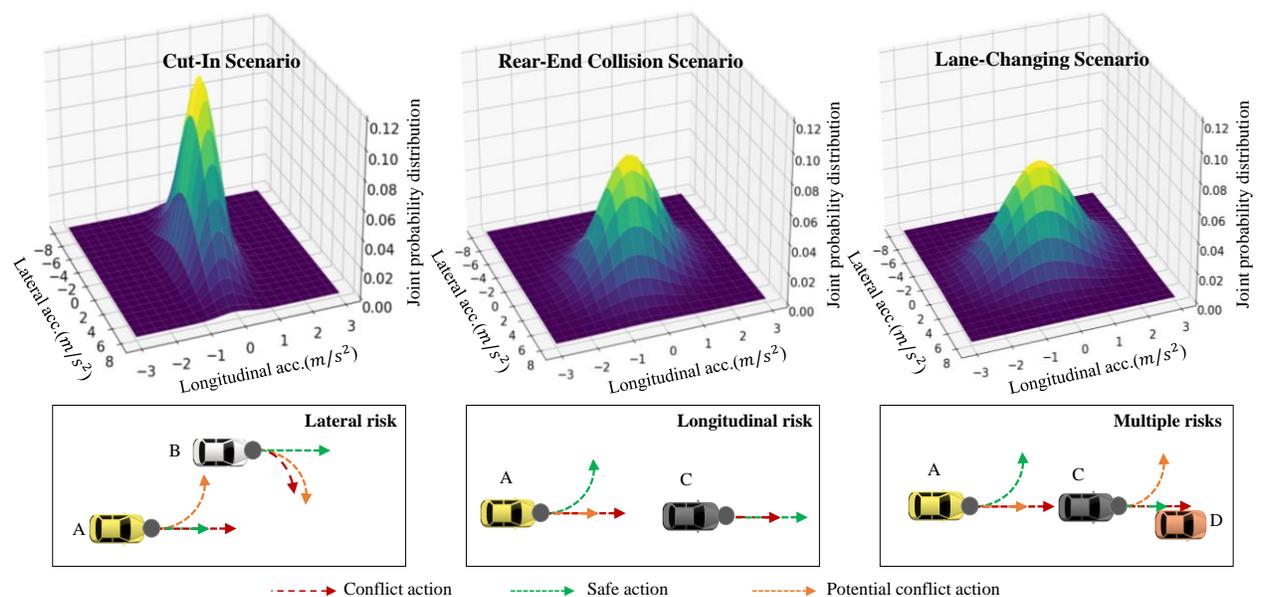

**Figure 5. Model performance analysis in different scenarios.**

## 4. Drift-diffusion decision-making modeling

Given the diverse risk sensitivities among drivers, it's crucial to explore the mechanisms shaping their decision-making behaviors in high-risk scenarios with multiple risk sources (Huang et al., 2024). We introduce the DDM to simulate and articulate these behaviors. It quantifies how the brain accumulates uncertain information until a decision threshold is reached, integrating cognitive psychology and neuroscience to explain driver decision-making in complex scenarios (Ratcliff et al., 2016). Analyzing our dataset with DDM provides insights into participant behaviors and response times.

*4.1 Drift diffusion model*

In Fig. 6, decision strategies 1 and 2 are modeled with baseline outputs at thresholds C and -C, symbolizing the driver's decision criteria. Absent prior knowledge, the initial bias starts at zero, but it shifts if influenced by pre-existing knowledge. The drift rate affects how quickly drivers respond to risks and decide to brake, and the reaction time encompasses the decision-making duration. The DDM mathematically captures the psychological dynamics and various factors influencing driver decision strategies, detailing how choices unfold across different strategies. DDM provides a comprehensive framework that traces driver behavior from initial attention through to cognitive processing and final decision-making, effectively modeling the information accumulation in driver risk assessment and reaction.

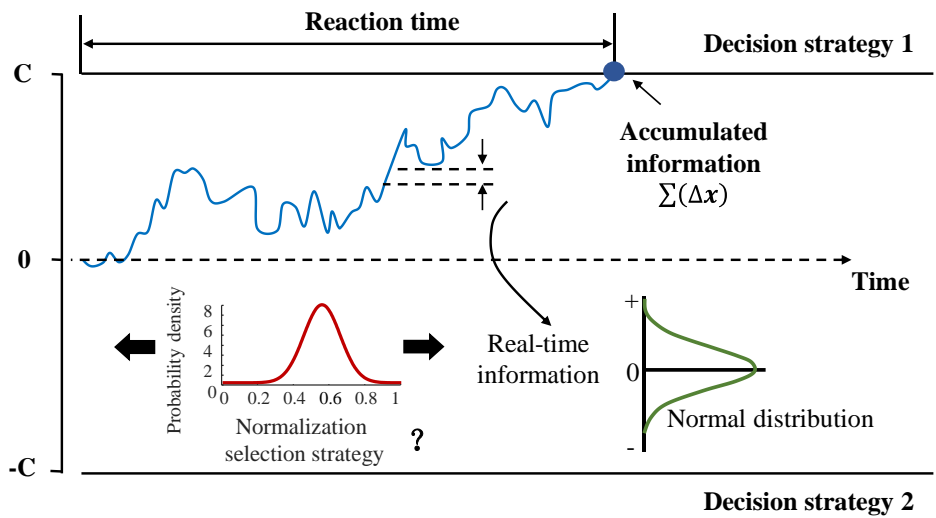

**Figure 6. The illustration of the drift-diffusion model.**

In the DDM, each strategy has a defined threshold dictating the necessary information accumulation for a response. Driver uncertainty introduces variability, influencing the accumulation direction. This model shows how decisions develop from information gathered over time. Illustrated in Fig. 6, DDM uses the search phase data and cognitive reaction times to time the initiation of avoidance actions in decision-making. This study applies DDM to analyze decision-making in three typical high-risk scenarios. Specifically, we will focus on the process that starts when B or C make interactions ($t = 0$), and ends when A decides to either steer or brake. Key components of the DDM in this analysis include drift rate, boundary settings, initial bias, and non-decision time.

*1) Drift rate*

The drift rate $g(t)$ characterizes the average rate of evidence accumulation over time, capturing both the speed and direction of the decision-making process. In this study, $g(t)$ is modeled as a function of the initial speed of the ego vehicle A, and the time headway as well as the distance between the ego vehicle and other surrounding vehicles and conflict vehicles. Referencing similar studies, we formulate the following drift rate expressions for the three scenarios examined in this study:

$$h^{AB}(t) = \frac{s^{AB}(t)}{v_0^A} \tag{5}$$

$$h^{AC}(t) = \frac{s^{AC}(t)}{v_0^A} \tag{6}$$

$$h^{AD}(t) = \frac{s^{AD}(t)}{v_0^A} \tag{7}$$

$$g(t) = \alpha(h^{AB}(t) + \kappa s^{AB}(t) + \gamma v_0^A - \theta) \tag{8}$$

$$g(t) = \alpha(h^{AB}(t) + \beta s^{AB}(t) + \delta h^{AC}(t) + \kappa s^{AC}(t) + \gamma v_0^A - \theta) \tag{9}$$

$$g(t) = \alpha(h^{AB}(t) + \beta s^{AB}(t) + \delta h^{AD}(t) + \kappa s^{AD}(t) + \gamma v_0^A - \theta) \tag{10}$$

Eqs. (5)-(7) define the calculations for time headways between the ego vehicle A and other surrounding vehicles, conflict vehicles, and stationary vehicles, $h^{AB}(t), h^{AC}(t), h^{AD}(t)$. The time headways are calculated by dividing the bumper-to-bumper distances, $s^{AB}, s^{AC}, s^{AD}$, by the speed of the ego vehicle $v_0^A$.

Eq. (8) defines the calculation for the drift rate in the cut-in scenario. The time headway between ego vehicle A and conflict vehicle B, $h^{AB}$, the distance between ego vehicle A and conflict vehicle B, $s^{AB}(t)$, and the initial speed of the ego vehicle A, $v_0^A$, are used in this calculation. $\alpha, \kappa, \gamma, \theta$ are free parameters to be calibrated.

Eq. (9) defines the calculation for the drift rate in the rear-end collision scenario. The time headway between ego vehicle A and conflict vehicle C, $h^{AC}$, the distance between ego vehicle A and conflict vehicle C, $s^{AC}(t)$, together with $h^{AB}$, $s^{AB}(t)$, and $v_0^A$ are used in this calculation. $\alpha, \beta, \delta, \kappa, \gamma, \theta$ are free parameters to calibrate.

Eq. (10) defines the calculation for the drift rate in the lane-changing scenario. The time headway between ego vehicle A and static vehicle D, $h^{AD}$, the distance between ego vehicle A and static vehicle D, $s^{AD}(t)$, together with $h^{AB}$, $s^{AB}(t)$, $v_0^A$, are used in this calculation. $\alpha, \beta, \delta, \kappa, \gamma, \theta$ are free parameters to calibrate.

*2) Boundary*

The boundary $b(t)$ defines the threshold of evidence required for decision-making. A decision is reached when the accumulated evidence meets either the upper or lower boundary, favoring the corresponding choice. Similar to the drift rate formulation, $b(t)$ is modeled as a function of the initial speed of the ego vehicle A, and the time headway as well as the distance between ego vehicle A and other surrounding vehicles and conflict vehicles, but utilizes the SoftMax function.

$$b(t) = \pm \frac{b_0}{1 + e^{-k(h^{AB}(t) + \kappa s^{AB}(t) + \gamma v_0^A - \tau)}} \tag{11}$$

$$b(t) = \pm \frac{b_0}{1 + e^{-k(h^{AB}(t) + \beta s^{AB}(t) + \delta h^{AC}(t) + \kappa s^{AC}(t) + \gamma v_0^A - \tau)}} \quad (12)$$

$$b(t) = \pm \frac{b_0}{1 + e^{-k(h^{AB}(t) + \beta s^{AB}(t) + \delta h^{AD}(t) + \kappa s^{AD}(t) + \gamma v_0^A - \theta - \tau)}} \quad (13)$$

Eq. (11) defines the calculation for the boundary in the cut-in scenario. $h^{AB}$, $s^{AB}(t)$, and $v_0^A$ are used in this calculation. $b_0$, $\kappa$, $\gamma$, $\tau$ are free parameters to be calibrated.

Eq. (12) defines the calculation for the boundary in the rear-end collision scenario. $h^{AB}$, $s^{AB}(t)$, $h^{AC}$, $s^{AC}(t)$, and A $v_0^A$ are used in this calculation. $b_0$, $\beta$, $\delta$, $\kappa$, $\gamma$, $\tau$ are free parameters to calibrate.

Eq. (13) defines the calculation for the boundary in the lane-changing scenario. $h^{AB}$, $s^{AB}(t)$, $h^{AD}$, $s^{AD}(t)$, and $v_0^A$ are used in this calculation. $b_0$, $\beta$, $\delta$, $\kappa$, $\gamma$, $\tau$ are free parameters to calibrate.

*3) Initial bias*

The initial bias, denoted as $Z$ represents the starting point of the evidence accumulation process. A negative value of $Z$ indicates an initial bias towards the "Brake" decision, while a positive value suggests a bias towards the "Steer" decision. In this study, we consider a SoftMax function of the initial speed of ego vehicle A, $v_0^A$, for the calculation of the initial bias for all three scenarios (Eq. (14)). $b_0$, $b_z$, $v$ are free parameters.

$$Z = \frac{2b_0}{1 + e^{-b_z(v_0 - v)}} - b_0 \quad (14)$$

*4) Non-decision time*

The non-decision time $t^{ND}$ represents the time taken by processes that are not directly related to the decision-making process itself. These processes include stimulus encoding, motor response execution, and any other processes that occur before or after the actual evidence accumulation. In this study, we assume a Gaussian distributed non-decision time shown in Eq. (15) for all three scenarios. $\mu^{ND}$ and $\sigma^{ND}$ are free parameters to be estimated.

$$t^{ND} \sim N(\mu^{ND}, \sigma^{ND}) \quad (15)$$

*5) Drift diffusion model formulation*

The formulation of the DDM is shown in Eq. (16), where $x(t)$ represents the evidence at time $t$, Positive values of $x(t)$ support the decision to "Steer", while negative values favor the decision to "Brake". $g(t)$ is the drift rate for the evidence accumulation defined in Eqs. Eq. (8)-(10). $\varepsilon(t)$ is the random noise added to the evidence. The drift rate, boundary, initial bias and non-decision time for cut-in scenario are shown in Eq. (8), (11), (14), (15) . for rear-end collision scenario are shown in Eq. (9), (12), (14), (15), for lane-changine scenario are shown in Eq. (10), (13), (14), (15).

The formulation of the DDM is shown in Eq. (16), where $x(t)$ represents the evidence at time $t$. Positive values of $x(t)$ support the decision to "Steer", while negative values favor the decision to "Brake". $g(t)$ is the drift rate for the evidence accumulation defined in Eqs. (8)-(10). $\varepsilon(t)$ is the random noise added to the evidence. The drift rate, boundary, initial bias, and non-decision time for the cut-in scenario are shown in Eqs. (8), (11), (14), and (15). For the rear-end collision scenario, they are shown in Eqs. (9), (12), (14), and (15), and for the lane-changing scenario, they are shown in Eqs. (10), (13), (14), and (15).

$$\frac{dx(t)}{dt} = g(t) + \varepsilon(t) \quad (16)$$

*4.2 Impact of risk sensitivity on DDM parameters*

The driver risk sensitivity model generates sub-models $R_s=\{R_{s,h}, R_{s,m}, R_{s,l}\}$, representing high, medium, and low sensitivity levels, quantifying individual differences in risk cognition. These sub-models directly influence key DDM parameters, adapting decision-making dynamics.

The drift rate $g(t)$, representing decision speed, increases with risk sensitivity. High-risk-sensitive drivers accumulate information faster, leading to quicker decisions, while low-risk-sensitive drivers have a slower drift rate, indicating more cautious decision-making. The decision boundary $b(t)$, defining the required evidence for a decision, is lower for high-risk-sensitive drivers, allowing them to decide with less accumulated information, whereas low-risk-sensitive drivers require more certainty. The initial bias $Z$ reflects decision inclination. Higher risk sensitivity shifts preference toward lane-changing, while lower sensitivity favors braking:

$$g'(t) = g(t) + \lambda R_s, \; b'(t) = b(t)e^{-\eta R_s}, \; Z'(t) = Z + \rho R_s, \tag{17}$$

where $\lambda$ is a scaling factor, $\eta$ controls the degree of adjustment based on risk sensitivity, $\rho$ is a tunable parameter determining the effect of risk sensitivity on initial bias.

By integrating risk sensitivity into DDM, the model dynamically adjusts decision-making speed, certainty thresholds, and action preferences, providing a more realistic and adaptive simulation of driver behavior in high-risk scenarios.

## 5. Results

In this section, we analyze the modeling results in detail. First, we identify the free parameters for each DDM and calibrate them to ensure optimal model performance. The accuracy of the calibrated DDM is then evaluated to assess its reliability in predicting driver decision-making. Finally, leveraging the well-fitted models, we interpret drivers' decision-making processes across the three scenarios from a cognitive perspective, providing insights into how risk perception and information accumulation influence their behavioral responses.

*5.1 Model parameters determining*

We implement the DDMs using the pyddm library and calibrate them to the actual data using a differential evolution optimization algorithm. The Bayesian Information Criterion (BIC) was employed as the loss function for this optimization process. **Table 3** presents the calibration results of the DDMs for the cut-in scenario, rear-end collision scenario, and lane-changing scenario.

**Table 3.** Parameters calibration results for the DDM in three scenarios.

| Scenario | $\alpha$ | $\beta$ | $\delta$ | $\kappa$ | $\gamma$ | $\theta$ | $b_0$ | $k$ | $\tau$ | $\mu^{ND}$ | $\sigma^{ND}$ | $b_z$ | $\nu$ |
|---|---|---|---|---|---|---|---|---|---|---|---|---|---|
| Cut-in | 0.07 | - | - | 0 | 0.63 | 71.97 | 0.56 | 1.83 | 1.50 | 1.33 | 0.23 | 0.12 | 4.07 |
| Rear-end collision | 1.09 | 0.57 | 0.00 | 1.00 | 1.59 | 79.75 | 0.60 | 0.95 | 5.43 | 0.61 | 0.17 | 0.09 | 14.71 |
| Lane-changing | 0.04 | 0.70 | 0.27 | 0.34 | 0.52 | 23.49 | 0.50 | 0.02 | 3.34 | 0.92 | 0.24 | 0.02 | 14.93 |

The accuracy of the DDM is validated by comparing cumulative response time probabilities between actual data and predictions from the established DDM. Fig. 7 shows a comparison example from the rear-end collision scenario. In the actual data, for both steer and brake decisions, the response times generally decrease with increasing initial speed of ego vehicle A, $v_0^A$. The response times for $v_0^A$ of 19.56 m/s and 22.10 m/s are similar, and the response times for $v_0^A$ of 23.32 m/s and 25.80 m/s are also similar. When $v_0^A$ is lower, there is no significant difference in response times between steer and brake decisions. However, when the initial speed is higher, the response time for brake decisions is notably shorter than for steer decisions. Comparing the remaining small graphs, it can be observed that the trend in the cumulative probability shown by the established DDM is similar to the trends in the actual data. This indicates that the established model accurately represents the original data with high precision.

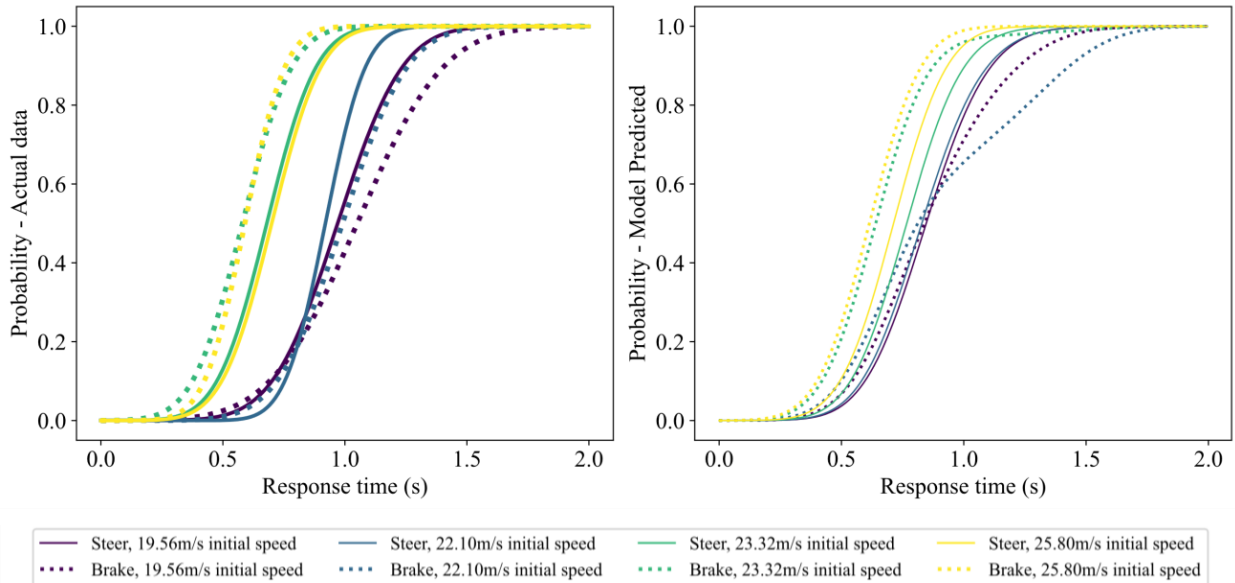

**Figure 7. Comparison of cumulative response time probabilities between actual data and predictions from the established DDM in the rear-end collision scenario.**

*5.2 Generalized decision-making process*

We now interpret the driver's decision-making process from a cognitive perspective. The established DDM will be used to simulate decision-making in the cut-in scenario, rear-end collision scenario, and lane-changing scenario.

*1) Cut-in scenario*

Our objective is to evaluate whether our models effectively captured the general trends in participant behavior, rather than explaining individual differences, and to use these models to interpret the general trends of the decision-making process. Thus, we categorize the actual data into four groups based on the initial speed value of the ego vehicle A, $v_0^A$, with median initial speeds of 25.82 m/s, 29.39 m/s, 31.69 m/s, and 33.85 m/s for each group respectively in cut-in scenario. We then configure the DDM simulation conditions to correspond with these four data groups. Given that the DDM model incorporates random factors, such as $\varepsilon(t)$ and $t^{ND}$, we conduct 1000 simulations for each initial speed group to mitigate the influence of these random factors. The simulation results are summarized in Fig. 8.

From the left column of Fig. 8, it could be observed that when faced with a sudden cut-in by a surrounding vehicle, drivers typically choose to brake directly rather than steer. The observed data reveals that when the ego vehicle's speed $v_0^A$ is relatively low, 100% of drivers chose to brake immediately. At higher ego vehicle speeds, approximately 90% of drivers chose to brake, while about 10% opted to steer, presumably due to insufficient time to brake effectively. This trend is also reflected in the established DDM. The developed DDM model predicts that in cut-in scenarios, 95% of drivers would choose to brake directly.

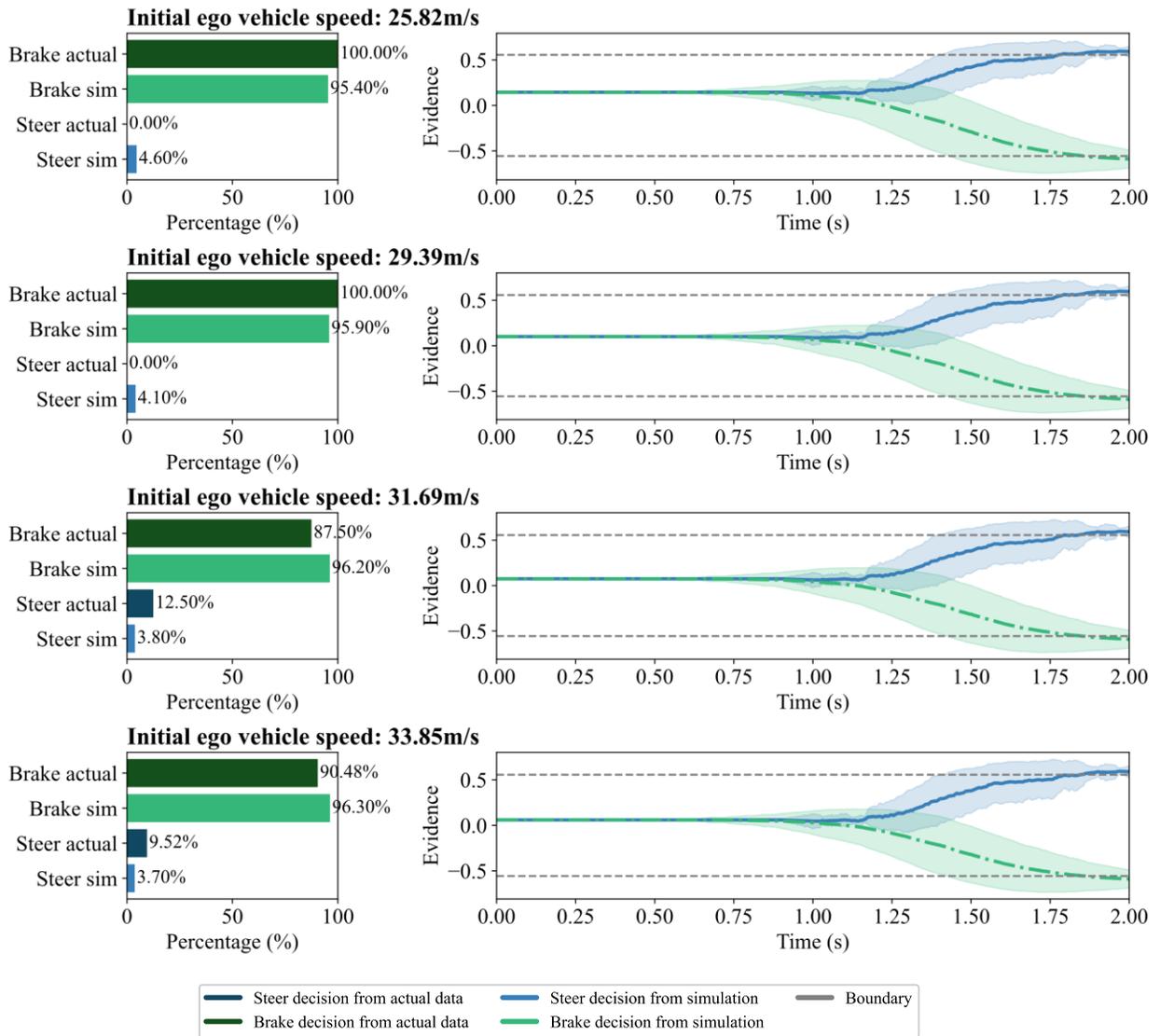

**Figure 8. Drivers' decision-making analysis in the cut-in scenario at different initial speeds from a cognitive perspective using the predicted DDM.** The left column shows the comparison of brake and steer decision ratios between actual data and DDM-predicted results. The right column illustrates the information accumulation process for both brake and steer decisions. The blue and green lines represent the accumulation process of evidence $x(t)$ for steer and brake, respectively. The grey dashed lines represent the boundaries that trigger steer and brake decisions. When the evidence $x(t)$ accumulates to either the steer or brake boundary, the driver will make the corresponding decision.

From the right column of Fig. 8, it can be seen that the accumulation of evidence to support the decision-making goes through two stages. (i) Non-decision period: Receiving information about the sudden cut-in by a surrounding vehicle, during which the evidence does not change over time. From the fitting results of $t^{ND}$, it is known that this process takes about 1.33s. Given that $Z > 0$, drivers may initially be inclined towards steering rather than braking when faced with this situation.

(ii) Evidence accumulation period: The driver begins to process information about the sudden cut-in by a surrounding vehicle and determines whether to steer or brake in response. The decision-making time for drivers doesn't vary significantly across different speeds. Although drivers initially lean towards steering, the value of the evidence $x$ supporting the decision rapidly decreases, quickly reaching the threshold that triggers the braking decision. From a cognitive perspective, this can be interpreted as follows: When confronted with a sudden cut-in, drivers initially consider steering as a potential evasive action. However, as they rapidly assess the situation, the accumulation of evidence strongly favors braking.

*2) Rear-end collision scenario*

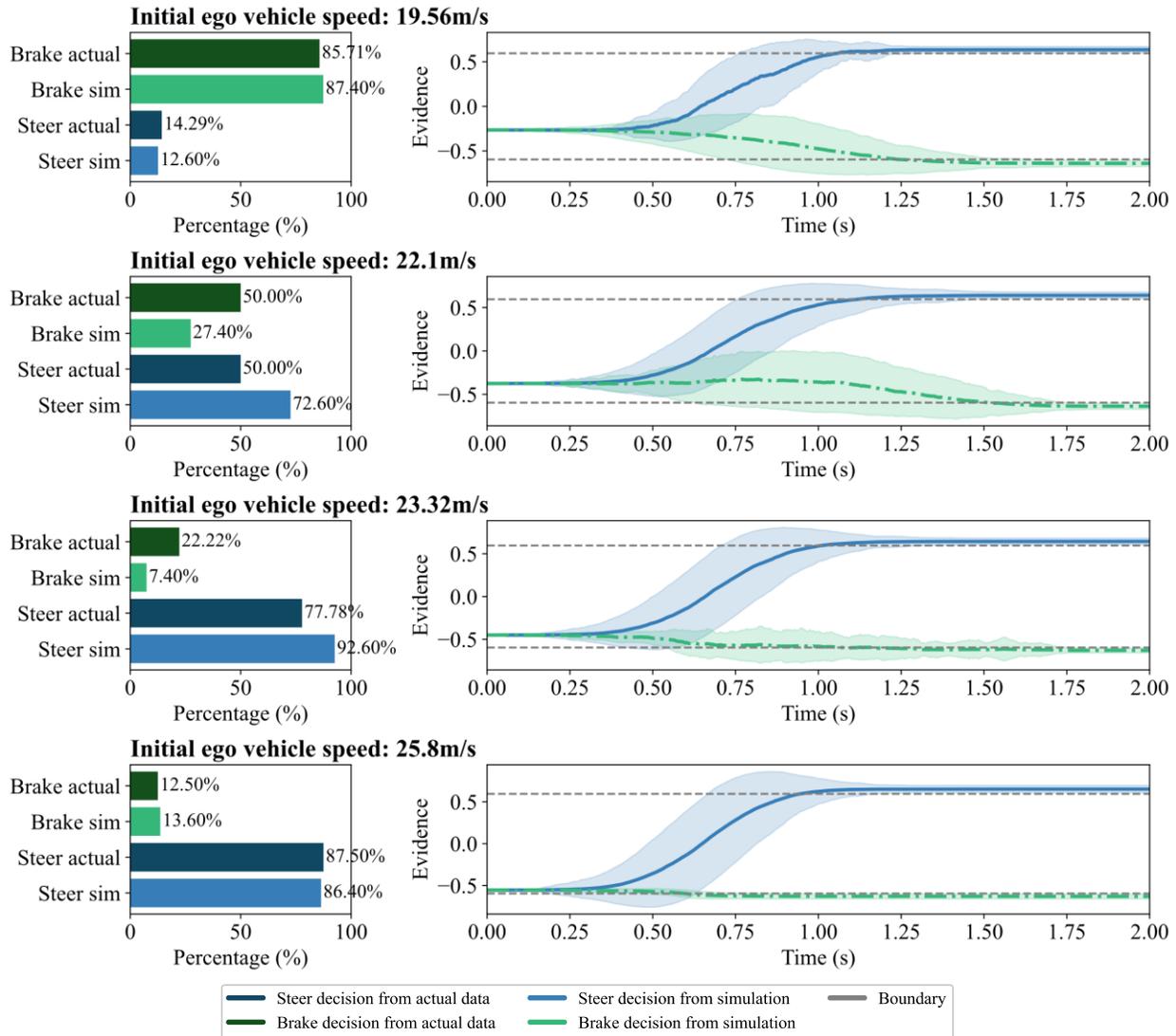

**Figure 9. Drivers' decision-making analysis in the rear-end collision scenario at different initial speeds from a cognitive perspective using the predicted DDM.**

The actual data is grouped into four groups based on the initial speed value of the ego vehicle A, $v_0^A$, with median initial speeds of 19.56 m/s, 22.10 m/s, 23.32 m/s, and 25.80 m/s for each group respectively in rear-end collision scenario. The DDM simulation conditions are configured to correspond with these four data groups. 1000 simulations for each initial speed group are conducted to mitigate the influence of these random factors. The simulation results are shown in Fig. 9.

From the left column of Fig. 9, when the initial speed is relatively low, drivers tend to choose to make a brake decision when faced with an emergency brake by a leading vehicle. As the initial speed increases, the probability of choosing the brake decision gradually decreases, while the probability of choosing the steer decision increases. When the initial speed is relatively high, facing an emergency brake by a leading vehicle, the time required to reduce the speed of an ego vehicle through braking is longer. Therefore, drivers will not choose to brake but instead make an immediate steer decision to respond. It could also be found that the decision-making probabilities predicted by the established DDM are close to probabilities from the actual data.

From the right column of Fig. 9, the accumulation of evidence also goes through two stages when making decisions. (i) Non-decision period: Receiving information about the emergency brake of the leading vehicle, during which the evidence does not change over time. From the fitting results of $t^{ND}$, it is known that this process takes about 0.6s. (ii) Evidence accumulation period: When the initial speed is relatively low, the response time required for drivers to make steer and brake decisions is similar. As the speed increases, the response times for both steer and brake decisions decrease, but the decrease in brake response time is more drastic. At this time, the driver will first judge whether there is a suitable opportunity to reduce the speed by braking. If so, they will immediately make a brake decision. If not, the driver will choose an opportunity to make a steer decision.

*3) Lane-changing scenario*

In the rear-end collision scenario, the actual data is also categorized into four groups based on the initial speed $v_0^A$ of ego vehicle. These groups have median initial speeds of 20.71 m/s, 23.27 m/s, 24.62 m/s, and 27.46 m/s respectively. The DDM simulation conditions are configured to correspond with these four data groups. To mitigate the influence of random factors, 1000 simulations are conducted for each initial speed group. Simulation results are presented in Fig. 10.

From the left column of Fig. 10, when faced with a sudden lane change by the leading vehicle and the appearance of a stationary obstacle ahead, drivers' choices between braking and steering to avoid the obstacle are roughly evenly split. Moreover, the proportion of drivers choosing to brake versus those to steer shows no significant variation across different initial speeds of the ego vehicle. This trend is well captured by the established DDM.

From the right column of Fig. 10, the accumulation of evidence also goes through two stages as other two scenarios. (i) Non-decision period: During this initial stage, drivers receive visual and situational information regarding the presence of a stationary obstacle ahead. From the fitting results of $t^{ND}$, it is known that this process takes about 0.92s. Since that $Z < 0$, drivers may initially be inclined towards braking rather than steering, This suggests that braking is perceived as the more immediate and intuitive response before further evidence accumulation refines the decision-making process. (ii) Evidence accumulation period: Information supporting both braking and steering accumulates at similar rates. The accumulated evidence typically reaches the decision-triggering boundary between 1.25 and 1.5 seconds. Notably, this process shows no significant variation across different initial speeds of the ego vehicle. From a cognitive perspective, this pattern suggests that drivers may process information for both potential actions, braking and steering, concurrently and with similar efficiency. The consistent timing of decision boundary attainment, regardless of initial speed, indicates a relatively stable cognitive processing time for lane-changing scenarios.

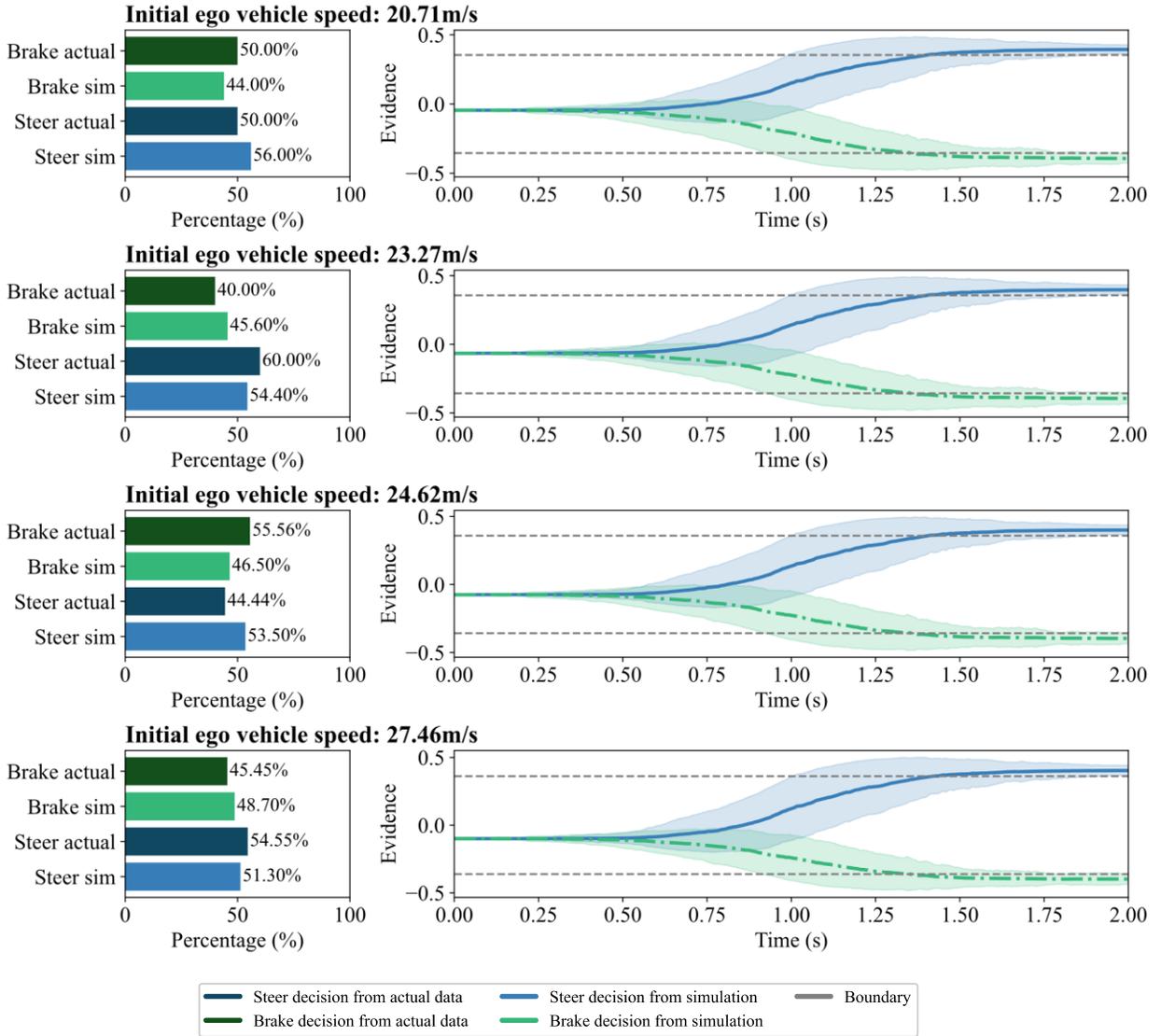

**Figure 10. Drivers' decision-making analysis in the lane-changing scenario at different initial speeds from a cognitive perspective using the predicted DDM.**

*5.3 Personalized decision-making process with risk sensitivity*

To further investigate the influence of driver risk sensitivity on information accumulation and decision-making, experiments were conducted across three high-risk scenarios: cut-in, rear-end collision, and lane change. The analysis focused on decision-triggering moments and collision avoidance behaviors among drivers with varying risk sensitivity levels, quantifying their impact on the evidence accumulation process.

As illustrated in Fig. 11, the black curve represents the evidence accumulation process, while the red dashed line denotes the fitted average decision threshold. The results indicate that risk sensitivity directly influences the drift rate (speed of evidence accumulation) and decision threshold (decision-triggering point), leading to distinct collision avoidance patterns.

Experimental findings show that driver risk sensitivity plays a crucial role in shaping the personalized decision-making process. High-risk sensitivity drivers ($R_{s,h}$) exhibit faster evidence accumulation and lower decision thresholds, allowing them to make quicker avoidance decisions and significantly reduce collision risk. In contrast, low-risk sensitivity drivers ($R_{s,l}$) demonstrate delayed responses in high-risk situations. For instance, in rear-end collision and lane-changing scenarios (Fig. 11 (b) and (c)), their decision-triggering moments occur above the threshold (Late Decision), requiring greater evidence accumulation before initiating action, making them more susceptible to collisions (red markers in Fig. 11). Compared to conventional driving models, the DDM integrated with risk sensitivity modeling enables a more accurate representation of individual decision-making tendencies, allowing for personalized, risk-aware decision adaptation AVs.

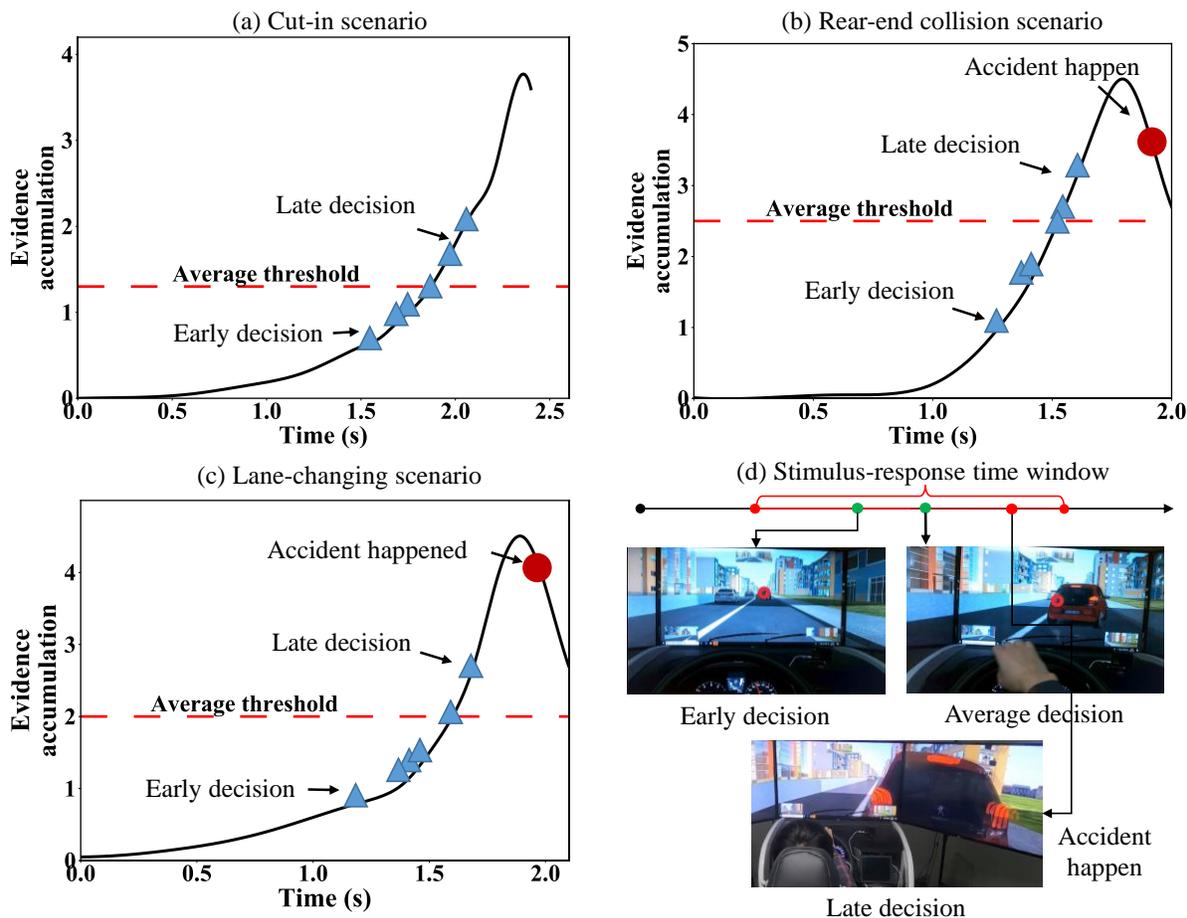

**Figure 11. Driver risk response and information accumulation process analysis.**

*5.4 Comparative experimental analysis*

To validate the advantages of the DDM framework for modeling driver decisions in high-risk scenarios, comparative experiments were conducted against classical driver models (IDM, MOBIL, and Gipps). The experiments focused specifically on the high-risk cut-in scenario, assessing model performance through decision accuracy and collision rate.

*1) Experimental setup and parameter calibration*

Experiments were conducted under identical high-risk scenarios as described previously (Section 6.2), with comparative analyses performed across four initial speed conditions (25.82 m/s, 29.39 m/s, 31.69 m/s, and 33.85 m/s). Parameters for IDM and MOBIL were calibrated according to standard values from prior studies (Kesting et al., 2007; Treiber et al., 2006), while Gipps model parameters were optimized based on safe-distance theory (Gipps, 1981). DDM parameters were fitted from experimental data via Bayesian optimization (Table 3). IDM and Gipps modeled only longitudinal decisions, and MOBIL integrated instantaneous lane-change decisions without cognitive accumulation. The DDM, however, incorporated dynamic information accumulation reflecting drivers' cognitive evaluation of risks.

Specifically, IDM adjusts acceleration based solely on spacing and relative speed without cumulative decision time. Gipps adjusts acceleration according to safe distance and acceptable speeds, assuming instantaneous decisions without information accumulation. MOBIL evaluates lane-changing decisions by maximizing lane-change benefits while accounting for safety and dynamic factors, but lacks cumulative decision time.

*2) Evaluation metrics and results analysis*

Decision accuracy measures the agreement percentage between model predictions (braking/steering) and observed driver behaviors, while collision rate indicates the proportion of simulation runs resulting in collisions. As shown in Table 4, at lower speeds (25.82 m/s and 29.39 m/s), DDM maintained a consistently high accuracy (>95%) with zero collision occurrences, comparable to other models. At higher speeds and elevated risks (31.69 m/s and 33.85 m/s), despite a slight decrease in accuracy (84.65% and 87.46%, respectively), DDM continued to achieve zero collisions, significantly outperforming MOBIL (collision rates of 2.7% and 1.4%) and IDM/Gipps (collision rates of 1.6% and 2.1%). IDM and Gipps exhibited perfect accuracy (100%) at lower speeds while showing reduced accuracy and increased collision rates at higher speeds due to their inability to model lane-change decisions.

**Table 4.** Parameters calibration results for the DDM in three scenarios.

| Initial speed (m/s) | Model | Decision accuracy (%) | Collision rate (%) |
|---|---|---|---|
| 25.82 | DDM | 95.4 | 0 |
| | MOBIL | 79.6 | 0 |
| | IDM/ Gipps | 100 | 0 |
| 29.39 | DDM | 95.9 | 0 |
| | MOBIL | 83.1 | 0 |
| | IDM/ Gipps | 100 | 0 |
| 31.69 | DDM | 84.65 | 0 |
| | MOBIL | 84.4 | 2.7 |
| | IDM | 87.5 | 1.6 |
| 33.85 | DDM | 87.46 | 0 |
| | MOBIL | 86.77 | 1.4 |
| | IDM/ Gipps | 90.5 | 2.1 |

DDM's superior collision avoidance performance can be attributed to its adaptive drift rates and decision boundaries, dynamically adjusted based on driver risk perception. In contrast, MOBIL and IDM/Gipps rely on fixed thresholds, limiting their responsiveness and accuracy in detecting and addressing risks under challenging high-speed conditions.

Experimental results demonstrate that the DDM more accurately predicts driver decisions in cut-in scenarios, effectively capturing cognitive processes and dynamic decision-making under high-risk conditions. In contrast, IDM and Gipps, lacking lateral decision-making modules, ensure only safe braking but cannot generate lane-change maneuvers. Although MOBIL incorporates lane-changing, it relies solely on instantaneous safety gaps without modeling drivers' cognitive accumulation and biases, resulting in notable deviations from observed behaviors. By dynamically adjusting thresholds through drift rates and decision boundaries, the DDM accurately captures driver behavior variations across speeds and risk levels. This finding highlights the importance of integrating cognitive dynamics into autonomous driving interaction design to enhance human behavior prediction.

# 6. Conclusion

This paper presents a cognition-decision framework that integrates risk sensitivity modeling and cognitive decision-making to enhance the understanding of driver behavior in high-risk scenarios. The risk sensitivity model, based on a multivariate Gaussian distribution, quantifies individual differences in risk cognition, capturing variations in how drivers perceive and respond to traffic risks. The DDM simulates decision-making by dynamically adjusting drift rate, boundary parameters, and initial bias based on driver-specific risk sensitivity, speed, and relative distance to other vehicles. Experimental validation in a driving simulator demonstrates that the proposed framework accurately predicts driver responses in emergency scenarios involving lateral, longitudinal, and multidimensional risk sources. Comparative analysis with IDM, Gipps, and MOBIL highlights the advantages of the DDM model in capturing cognitive decision processes and adaptive driving behaviors, particularly in scenarios requiring complex risk assessment and rapid decision-making. The results confirm the model's superior predictive accuracy and practical applicability in understanding human driver behavior, which is essential for improving AV-human interaction. The findings offer theoretical support for human-centered autonomous driving, enabling safer and more adaptive AV integration into mixed traffic. Future work will refine the model by incorporating real-time environmental perception and neurocognitive markers, enhancing AV decision-making for context-aware and human-compatible driving behaviors.

**CRediT authorship contribution statement**
Zheng Li: Methodology, software development, data processing, original draft preparation. Heye Huang: Methodology refinement, data curation. Hao Cheng: Conceptualization, draft revision for clarity. Haoran Wang/Junkai Jiang: Data visualization, validation of findings. Xiaopeng Li: Project conceptualization, supervision, review, and editing of the manuscript. Arkady Zgonnikov: Draft revision, enhancing content quality.

**Declaration of competing interest**
The authors declare that they have no known competing financial interests or personal relationships that could have appeared to influence the work reported in this paper.

**Acknowledgments**
This work was sponsored by the Center for Connected And Automated Transportation (CCAT) project "Traffic Control based on CARMA platform for maximal traffic mobility and safety", and also supported by the U.S. National Science Foundation (NSF) under Grant No.2313578. Furthermore, we thank Samir H.A. Mohammad for providing the source code used as a basis for model simulations in this work.